\documentclass[11pt]{article}

\usepackage[final]{acl}

\usepackage{times}
\usepackage{latexsym}
\usepackage[T1]{fontenc}
\usepackage[utf8]{inputenc}
\usepackage{microtype}
\usepackage{inconsolata}

\usepackage{amsmath}
\usepackage{amssymb}
\usepackage{booktabs}
\usepackage{multirow}
\usepackage{graphicx}          % includegraphics, resizebox
\usepackage[table,dvipsnames]{xcolor}
\usepackage{makecell}
\usepackage{enumitem}
\usepackage{pifont}
\usepackage{bbding}
\usepackage{wasysym}
\usepackage{tcolorbox}
\usepackage{threeparttable}
\usepackage{algorithm}
\usepackage{algorithmicx}
\usepackage{algpseudocode}
\usepackage{seqsplit}
\usepackage{fvextra}
\usepackage{xurl}              % better URL line breaking (loads url)

\definecolor{darkblue}{rgb}{0, 0, 0.5}
\hypersetup{colorlinks=true, citecolor=darkblue, linkcolor=darkblue, urlcolor=darkblue, breaklinks=true}

\definecolor{headerblue}{RGB}{219,234,254}
\definecolor{oursgreen}{RGB}{220,252,231}
\definecolor{oursbanner}{RGB}{187,247,208}
\definecolor{rowgray}{RGB}{248,248,248}
\definecolor{secbanner}{RGB}{243,244,246}

\setlist[itemize,enumerate]{leftmargin=*}

\title{When Only the Final Text Survives: Implicit Execution Tracing for Multi-Agent Auditing}

\author{
 \textbf{Yi Nian\textsuperscript{1,*}},
 \textbf{Haosen Cao\textsuperscript{1,*}},
 \textbf{Shenzhe Zhu\textsuperscript{2}},
 \textbf{Henry Peng Zou\textsuperscript{3}},
\\
 \textbf{Qingqing Luan\textsuperscript{4}},
 \textbf{Yudi Zhang\textsuperscript{4}},
 \textbf{Yue Zhao\textsuperscript{1,\dag}}
\\
\\
 \textsuperscript{1}University of Southern California,
 \textsuperscript{2}University of Toronto,
\\
 \textsuperscript{3}University of Illinois Chicago,
 \textsuperscript{4}Independent Researcher
\\
}

\begin{document}
\maketitle

\begin{abstract}
When a multi-agent system produces an incorrect or harmful answer, who is accountable if execution logs and agent identifiers are unavailable? In practice, generated content is often detached from its execution environment due to privacy or system boundaries, leaving the final text as the only auditable artifact. Existing attribution methods rely on full execution traces and thus become ineffective in such metadata-deprived settings. We propose \textbf{Implicit Execution Tracing (IET)}, a provenance-by-design framework that shifts attribution from post-hoc inference to built-in instrumentation. Rather than inferring
provenance after the fact, IET embeds agent-specific, key-conditioned
statistical signals into the token generation process at \emph{generation
time}, turning the output text into a self-verifying provenance record.
An \emph{offline auditor}, holding a verification registry that maps each agent
to its key, then recovers segment-level provenance---segment boundaries and
per-segment agent attribution---from the final text alone without access to execution logs or private traces. Experiments across diverse multi-agent coordination settings demonstrate that IET achieves accurate segment-level attribution and reliable transition recovery under identity removal, boundary corruption, and privacy-preserving redaction, while maintaining generation quality. These results show that embedding provenance into generation provides a
practical foundation for accountability in multi-agent language systems under metadata loss.
\end{abstract}

\section{Introduction}

The adoption of autonomous agents is increasing rapidly; industry forecasts
indicate that more enterprise applications will feature task-specific AI agents
by 2026 \citep{gartner2025enterpriseagents}. Despite
this growth, recent evaluations of multi-agent frameworks report substantial
failure in complex tasks \citep{cemri2025multifail,miao2025recodeh}. This
operational opacity creates a \textit{gap in accountability} when systems
produce incorrect or harmful content.

\paragraph{Background.}
As multi-agent systems take on tasks, their outputs must remain
\emph{auditable}: a party that was not present at execution should be able to
reconstruct what happened from evidence it can check for itself
\citep{nian2026auditable}. Auditing such a system begins with attribution,
since no question of responsibility can be settled before it is known which
agent produced which part of the output.
Attribution in language generation has relied on explicit metadata: model
identifiers, execution logs, or externally recorded provenance signals
\citep{wang2026trajectorysurvey, barke2026agentrx, zhang2025whichagent}. In
multi-agent language systems, this assumption becomes increasingly fragile.
Current state-of-the-art diagnostic frameworks, such as Who\&When
\citep{zhang2025whichagent} and FAMAS \citep{ge2025failureattributing}, require
analyzing complete execution trajectories to perform failure attribution. These
methods assume an optimal ``white-box'' environment where agent identities and
internal logic are fully transparent. However, exposing such detailed traces
introduces privacy risks and compliance constraints, especially when logs are
abstracted or redacted \citep{xiang2024guardagent}. In real-world production,
agent-generated content is frequently decoupled from its original environment
through simple ``copy-pasting'' into external reports or emails. Once text
leaves the server-side environment in this way, the link to its execution
metadata is severed and cannot be restored from the text alone---the same
observation that motivates carrying detection signals inside the text rather
than alongside it \citep{kirchenbauer2023watermark}. In scenarios where agent
identifiers are unavailable, interaction boundaries are not explicitly marked,
or sensitive information has been redacted, existing trajectory analysis tools
\citep{zhang2025whichagent,ge2025failureattributing} become ineffective,
leaving the generated text as the artifact available to an
auditor. This necessitates an intrinsic, self-verifying mechanism that allows provenance recovery and execution tracing directly from the output, without requiring access to original system traces. 

Furthermore, while prior work focuses predominantly on failure attribution within multi-agent systems \citep{wang2026trajectorysurvey,zhang2025whichagent,wu2025autonomousdrivingsurvey}, we argue that provenance recovery is a more fundamental problem. It is essential for auditor to understand "who said what" and "how they interacted", regardless of whether a failure occurred. In this sense, failure attribution can be viewed as a downstream task built upon reliable execution tracing.

\paragraph{Problem Definition.}
Multi-agent systems often involve complex and partially unobserved interaction
processes, while the observable output is a single linear text sequence. This
creates a gap between execution and observable output. We therefore formulate
the problem at the \emph{agent-segment level}: we model the text as contiguous
agent segments. The goal is to recover (1)~where segment boundaries occur, and
(2)~which agent each segment belongs to, using only the final output available
to the auditor. We recover the transition structure induced by the linearized output sequence,
the well-posed and audit-sufficient target under our ``only-the-final-text-survives''
setting.

\paragraph{The Audit Setting.}
IET supposes \emph{cooperative setting}: At \emph{generation time}, agents apply IET locally, each
holding its own key. At \emph{audit time}, a trusted \emph{auditor}
verifies provenance from the final text alone via a registry mapping each agent
to its key and tokenizer. Agents are assumed honest-but-instrumented, and no
participant needs another's logs, weights, or private traces: the setting
requires cooperation at generation time, not mutual disclosure. In this way,
provenance is fixed at the source while verification can occurs arbitrarily later.

\paragraph{Our Approach.}
We introduce IET, a keyed,
privacy-preserving auditing framework that encodes agent identity directly into
the token distribution during generation. Rather than relying on external logs
or metadata, IET injects statistically controlled, key-conditioned signals into
the decoding process, enabling recovery from the final text alone at audit
time. The auditor first detects segment boundaries using sliding-window
statistical scoring combined with change-point detection, and then assigns an
agent label to each recovered segment based on its aggregated attribution
signal. The resulting segment sequence induces a transition structure over
agents. This formulation supports robust boundary detection and segment-level
attribution across diverse coordination patterns, while remaining well-defined
even when the underlying execution is partially concurrent or unobserved.

\paragraph{Contributions.}
Our main contributions are summarized as follows:
\begin{itemize}[itemsep=1pt, topsep=2pt, parsep=0pt, partopsep=0pt]

\item \textbf{New Problem Formulation.}
We formalize \textit{segment-level provenance auditing} for multi-agent
language systems, where the goal is to recover segment boundaries and attribute
each segment to its originating agent from the final text alone, under metadata
loss. Unlike prior work on error or failure attribution, which assumes access
to execution logs, we treat provenance recovery itself as a first-class problem
under missing or obfuscated execution information.

\item \textbf{Keyed Implicit Tracing Mechanism.}
We introduce a keyed, distribution-level tracing mechanism that embeds
agent-specific signals into the token generation process. This enables
segment-level attribution directly from the final text at audit time, without
relying on explicit metadata or execution logs.

\item \textbf{Boundary Detection and Attribution.}
We develop a sliding-window statistical scoring method combined with
change-point detection to identify segment boundaries, and assign agent labels
to each recovered segment based on aggregated attribution signals. We further
represent the recovered transition structure as a transition graph.

\item \textbf{Empirical Evaluation.}
Across diverse multi-agent coordination scenarios, we demonstrate strong
recoverability under identity removal, boundary perturbation, and
privacy-preserving redaction, while preserving generation quality; we further
characterize the audit regime under which provenance remains verifiable.

\end{itemize}

\begin{figure*}[t]
    \centering
    \includegraphics[width=\textwidth]{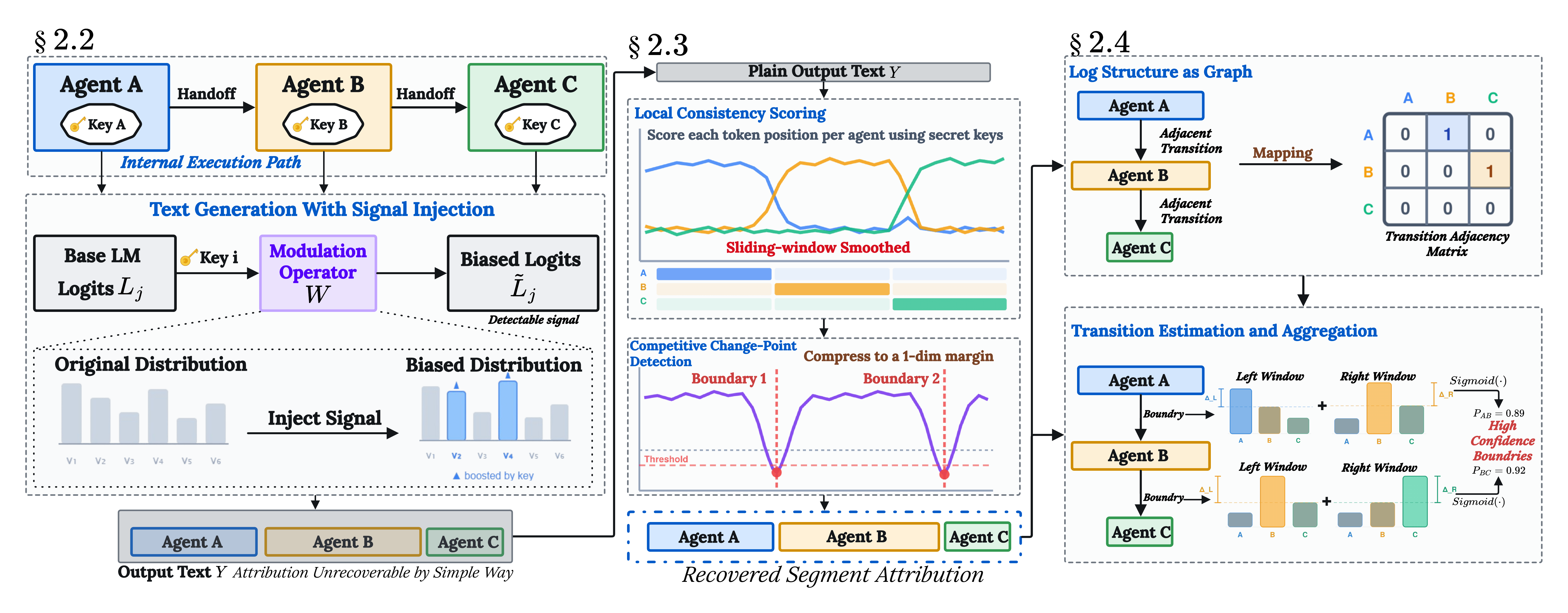}
    \caption{Overview.}
    \label{fig:Overview}
\end{figure*}

\section{Method}

\subsection{Problem Formulation}
\label{sec:problemformulation}
Given the final output sequence $Y = (y_1, y_2, \dots, y_T)$, our goal is to
recover a segmentation of the sequence into contiguous segments and assign each
segment to an agent $a \in \mathcal{A}$.
To this end, we define an attribution function $\hat{g}(t)$ over tokens, which
is constrained to be piecewise constant. Each constant region of $\hat{g}(t)$
corresponds to a segment associated with a single agent. Thus, while
attribution is computed at the token level, the objective is to recover
consistent segment-level boundaries and labels.

\paragraph{Attribution via Sequential Scoring.}
We assume a scoring function $f(t, a)$ that evaluates the alignment between the
context at time $t$ and agent $a$. The attribution $\hat{g}(t)$ remains assigned
to the current agent $a_k$ until a transition to a new agent
$a' \in \mathcal{A} \setminus \{a_k\}$ is identified by a detection operator
$D$:
\begin{equation}
    D(f, a_k, a', t, \mathcal{H}_t) > \tau.
\end{equation}
Here, $\mathcal{H}_t$ incorporates the temporal history of scores, and $\tau$
is a sensitivity threshold. This formulation uses token-level signals to detect
segment boundaries, where transitions are triggered by changes in relative
alignment scores between competing agents.

\paragraph{Robustness to Metadata Loss.}
To evaluate the restorative utility of our method, we consider a
\textbf{metadata obfuscation operator} $\mathcal{O}$ that simulates
metadata-deprived conditions by stripping agent identifiers and segment
boundaries. Let $\Phi(G, Y)$ be the performance of a downstream model $G$
(e.g., error attribution) given a fully annotated output sequence $Y$. Our
objective is to ensure \textit{attribution consistency}:
\begin{equation}
\Phi\!\left(G, \mathrm{Rec}(\mathcal{O}(Y), \hat{g})\right)
\approx
\Phi(G, Y)
\end{equation}

where $\mathrm{Rec}$ is the reconstruction function powered by our estimated
attribution $\hat{g}$. This formulation characterizes our method as a robust
backbone that maintains the diagnostic power of downstream tasks even under
severe metadata loss.

\paragraph{Transition Structure Validation.}
To evaluate the consistency of the recovered attribution, we construct a
transition graph $\hat{G} = (\mathcal{A}, \hat{\mathcal{E}})$ from the recovered
segment sequence.
An edge $(a^{(i)}, a^{(j)}) \in \hat{\mathcal{E}}$ is added whenever a segment
attributed to agent $a^{(i)}$ is followed by a segment attributed to agent
$a^{(j)}$, where $i \neq j$. This graph captures the transition structure
induced by the output sequence, rather than reconstructing the underlying
communication topology, which may be partially unobserved or concurrent.

% =====================================================================
%  IET — Revised Method (Embedding + Attribution + Transition Encoding)
%  ARR resubmission. Changes applied:
%   #1  Opening "post-hoc attribution" -> "verifiable attribution"
%       + "at generation time" (post-hoc is reserved for baselines).
%   #2  "identified in the execution trace" -> active agent known
%       locally at generation time (no trace dependency).
%   #3  All "log" wording removed/aligned:
%        - "Log Transition Structure Encoding" -> "Transition Structure Encoding"
%        - "Log Structure as Graph" -> "Transition Structure as Graph"
%        - "linearized execution log" -> "linearized output sequence"
%        - "interaction log" -> "output sequence Y"
%        - "log-level transition structure" -> "transition structure"
%   #4  Sequence symbol unified to Y (generated text) throughout.
%   #5  Segment count uses S (agent count stays K = |A|) to avoid clash.
%  KEPT unchanged: the two honesty statements that the transition
%  structure does NOT reconstruct the communication/execution topology.
%  NOTE: make sure no stray \mathcal{L} or $Y$-vs-$\mathcal{S}$ mismatch
%  remains elsewhere (experiments / appendix).
% =====================================================================

\subsection{Keyed Implicit Embedding}

To enable verifiable attribution in metadata-deprived environments, we bind
each agent's identity directly to its generated output through keyed modulation
of the token distribution at generation time. This process introduces
agent-specific statistical signatures that can later be verified using the
corresponding secret key, following the statistical signaling framework
introduced by \citet{lau2024waterfall}. Let $L_j \in \mathbb{R}^{|\mathcal V|}$
denote the logits produced by the base language model at token position $j$.
For the active agent $a_k \in \mathcal{A}$ generating at this step, we produce
text by applying a keyed distributional modulation operator $\mathcal{W}$
 on the agent identity:
\begin{equation}
\tilde L_j = \mathcal{W}(a_k, L_j),
\end{equation}
where $\mathcal{W}$ induces a statistically detectable bias determined by
$a_k$. Specifically, the agent identity $a_k$ induces two distinct keys:
\begin{equation}
k_p = h_p(a_k), \qquad k_\pi^{(j)} = h_\pi(a_k,\, y_{j-n+1:j-1}),
\label{eq:waterfall_keys}
\end{equation}
where $h_p$ and $h_\pi$ are deterministic, key-conditioned hash functions that
map the agent identity to pseudo-random vectors and permutations, respectively.
Therefore, $k_p$ determines a fixed perturbation direction in logit space
unique to agent $a_k$, and $k_\pi^{(j)}$ determines a context-dependent
vocabulary permutation. Here $y_{j-n+1:j-1}$ represents the length-$(n-1)$ context window used
to derive the context-dependent key. Similar to \citet{lau2024waterfall}, the
logits are then modulated as:
\begin{equation}
\tilde L_j = \mathcal{P}^{-1}\!\Big( k_\pi^{(j)},\; \mathcal{F}\!\big( k_p, \kappa, \mathcal{P}(k_\pi^{(j)}, L_j) \big) \Big),
\end{equation}
where $\mathcal{P}$ handles the vocabulary permutation and $\mathcal{F}$ applies
a low-magnitude perturbation $\kappa$ along the direction $k_p$. Sampling from
this modified distribution embeds the agent's identity $a_k$ as a persistent
statistical signal, enabling structural recovery even when explicit metadata is
removed.

\subsection{Agent Attribution}
IET estimates the attribution function $g(t)$ by identifying which agent's
signal best explains each segment of the generated text $Y$; we write
$\hat{g}(t)$ for the resulting estimate.

\paragraph{Local Consistency Scoring.}
We first define a token-level statistic $x_j(a)$ for the token at position $j$
to measure the alignment between the observed token $y_j$ and the signal of
agent $a \in \mathcal{A}$:
\begin{equation}
x_j(a)=k_p(a)\bigl[\mathcal{P}(k_\pi^{(j)}, y_j)\bigr].
\end{equation}
At each position $j$, we compute a score by applying the permutation
$\mathcal{P}(k_\pi^{(j)}, \cdot)$ to the observed token $y_j$ and evaluating it
under the agent-specific function $k_p(a)$.
To mitigate local variance, we instantiate the \textbf{agent-consistency
metric} $f(t, a)$ from our formulation using a sliding window of width $w$:
\begin{equation}
f(t, a) = \frac{1}{w} \sum_{j=t}^{t+w-1} x_j(a).
\end{equation}
This window-averaged score $f(t, a)$ bridges the gap between low-level token
statistics and high-level structural inference.

\paragraph{Competitive Change-Point Detection.}
To identify transitions between agents, we instantiate the detection operator
$D$ using a competitive detector. We first reduce the multi-agent consistency
scores into a one-dimensional competitive margin signal $z_t$, which quantifies
the relative dominance of the leading agent candidate:
\begin{equation}
z_t = f(t, \hat{a}_t) - \max_{a' \in \mathcal{A} \setminus \{\hat{a}_t\}} f(t, a'),
\end{equation}
where $\hat{a}_t = \arg\max_{a \in \mathcal{A}} f(t, a)$ is the agent with the
highest alignment score at time $t$.

To robustly localize transitions, we smooth the margin sequence $\{z_t\}$ with
a cumulative sum algorithm. For the $s$-th segment starting at position
$\hat{\tau}_{s-1}$, we track the cumulative deviation from the local mean margin
$\bar{z}_s$:
\begin{equation}
C_t = \sum_{i=\hat{\tau}_{s-1}}^{t} \bigl(z_i - \bar{z}_s\bigr).
\label{eq:recursive_cusum}
\end{equation}

We formally define the detection operator $D$ as a binary decision at the
potential change point $\hat{\tau}_s = \arg\max_{t > \hat{\tau}_{s-1}} |C_t|$:
\begin{equation}
D(f, a',\hat{a}_t ,\hat{\tau}_s) = \mathbb{I} \left( \lvert C_{\hat{\tau}_s} \rvert > \tau \right),
\label{eq:detector_definition}
\end{equation}
where $\tau$ is the sensitivity threshold. By iteratively identifying these
extrema, we partition the output sequence $Y$ into $S$ contiguous segments.
This induced segmentation allows us to recover the \textbf{attribution
function} $\hat{g}(t)$ as a piecewise-constant mapping:
\begin{equation}
\hat{g}(t) = \hat{a}_s, \quad \text{for } t \in [\hat{\tau}_{s-1}, \hat{\tau}_s),
\label{eq:attribution_function}
\end{equation}
where $\hat{a}_s$ is the dominant agent identified within the $s$-th segment.

\subsection{Transition Structure Encoding}

To capture interaction patterns from the recovered attribution, we encode the
transition structure induced by the segment sequence. We emphasize that this
transition structure is derived from the \textbf{linearized output sequence},
and does not necessarily coincide with the underlying agent communication
graph. Instead, it provides a sequential approximation of interaction flow as
observed in the generated text.

\paragraph{Transition Structure as Graph.}
Let $\mathcal{A} = \{a_1, \dots, a_K\}$ be the set of agents. At any step $t$,
we construct a binary adjacency matrix $\mathbf{M}_t \in \{0,1\}^{K \times K}$,
where $(\mathbf{M}_t)_{i,j} = 1$ indicates that a transition from agent $a_i$ to
agent $a_j$ has been observed in the recovered segment sequence.

The matrix $\mathbf{M}_t$ is incrementally accumulated as new segment
transitions are detected. We encode this transition pattern into a compact
identifier $\mu_t$ by flattening $\mathbf{M}_t$ into a bitmask:
\begin{equation}
\mu_t = g(\mathbf{M}_t) = \sum_{i=1}^{K} \sum_{j=1}^{K} (\mathbf{M}_t)_{i,j} \cdot 2^{(i-1)K + (j-1)}.
\label{eq:matrix_encoding}
\end{equation}
This encoding represents the set of observed pairwise transitions, allowing us
to distinguish transition patterns induced by the segment sequence, rather than
reconstructing the full underlying execution or communication topology.

\paragraph{Transition Estimation and Aggregation.}
For each detected change point $\hat\tau_s$, we consider its local neighborhood
as determined by the same sliding-window procedure used in boundary detection,
and compute agent-consistency scores over stable regions immediately before and
after:
\begin{equation}
\bar f_L(a)=\frac{1}{h}\sum_{t=\hat\tau_s-h}^{\hat\tau_s-1} f(t,a),
\bar f_R(a)=\frac{1}{h}\sum_{t=\hat\tau_s+1}^{\hat\tau_s+h} f(t,a),
\end{equation}
where $h$ is the width of the neighborhood. For a candidate transition $a_i \to a_j$, we define a
local transition confidence:
\begin{equation}
\begin{split}
P_{ij} = \sigma\Big(
&\big[\bar f_L(a_i)-\max_{k\neq i}\bar f_L(a_k)\big] \\
&+ \big[\bar f_R(a_j)-\max_{k\neq j}\bar f_R(a_k)\big]
\Big),
\end{split}
\end{equation}
where $\sigma(x)=1/(1+e^{-x})$ is the logistic function, so that $P_{ij}$ is
large only when $a_i$ dominates on the left and $a_j$ dominates on the right.
A log typically contains several boundaries; we aggregate them into a single
predicted adjacency matrix by taking, for each ordered pair,
\begin{equation}
\hat A_{ij} = \max_{s\in\mathcal{B}_{ij}} P_{ij}^{(s)},
\end{equation}
where $\mathcal{B}_{ij}$ is the set of detected boundaries whose adjacent
segments are attributed to $a_i$ and $a_j$, and $\hat A_{ij}=0$ when
$\mathcal{B}_{ij}=\varnothing$. This keeps $\hat A$ on the same support as the
ground-truth adjacency matrix. This formulation favors boundaries supported by
stable segment-level attribution, rather than isolated near-threshold
fluctuations.

\section{Experiment}

\subsection{Datasets}

We evaluate on two multi-agent benchmarks with ground-truth speaker
annotations; two reasoning benchmarks used only to measure generation quality
are described in Appendix~\ref{app:dataset_details}.

\noindent\textbf{Multi-Agent Interaction Dataset}~\citep{liu2025topologymatters}.
This dataset contains structured multi-agent dialogues under diverse coordination paradigms, including hierarchical delegation and collaborative planning. 
Each interaction log provides explicit agent identities and turn-level boundaries, allowing analysis of agent participation and transition patterns. Further details of the MAMA topology dataset are provided in Appendix~\ref{app:mama_topology_dataset}.

\noindent\textbf{Who \& When}~\citep{zhang2025whichagent}.
This benchmark focuses on speaker and error attribution in multi-agent transcripts. 
The dataset contains conversational logs with annotated speaker identities and temporal ordering, enabling evaluation of attribution in multi-party interactions. Additional details of the Who\&When benchmark are provided in Appendix~\ref{app:who_when_dataset}.

\subsection{Experimental Design}

To evaluate the capabilities defined in Section~\ref{sec:problemformulation}, we design two complementary experimental settings that test (i) structure-aware attribution under controlled interaction topologies and (ii) robustness of attribution under metadata obfuscation.

\paragraph{Structure-aware Attribution.}
We evaluate attribution and structural recovery using the Multi-Agent Interaction Dataset~\citep{liu2025topologymatters}. 
This dataset provides structured interaction logs generated under predefined coordination patterns among agents.
To analyze how interaction structure affects attribution difficulty, we consider several canonical coordination settings, including \emph{star}, \emph{chain}, and \emph{tree} configurations. 
For each setting, agents follow a fixed interaction protocol while solving the same set of tasks, resulting in logs with distinct transition patterns. These logs allow us to evaluate whether the recovered attribution can correctly identify agent contributions and capture the transition structure induced by the observed text.

\paragraph{Robust attribution under metadata loss.}
We evaluate robustness using the Who\&When benchmark~\citep{zhang2025whichagent}, which contains multi-agent transcripts with annotated speaker identities and error attribution labels. 
To simulate metadata-independent scenarios and verify IET framework's robustness, we apply two types of transcript perturbations:
(i) \textbf{ID Removal}, where explicit agent identifiers are removed from each utterance, and 
(ii) \textbf{Boundary Corruption}, where utterance boundaries are shuffled or partially merged.
These perturbations mimic realistic situations where execution metadata is unavailable or unreliable.
The goal is to evaluate whether agent identities and interaction structures can be recovered from text alone under increasing levels of transcript corruption.

\subsection{Evaluation Metrics}
\label{sec:metrics}
Following the problem formulation in Section~\ref{sec:problemformulation}, 
we evaluate three aspects of execution tracing: sequential attribution, 
structural validation, and robustness under metadata loss.

\paragraph{Sequential Attribution.}
We measure whether the recovered attribution function $\hat{g}(t)$ correctly
assigns tokens to generating agents. Let $\mathcal{T}_a=\{t: g(t)=a\}$ and
$\hat{\mathcal{T}}_a=\{t: \hat{g}(t)=a\}$ denote the token positions assigned to
agent $a$ under the ground truth and under our estimate. We report token-level
accuracy and the mean per-agent IoU:
\begin{equation}
\mathrm{TokenAcc} = \frac{1}{T}\sum_{t=1}^{T} \mathbb{I}[\hat{g}(t)=g(t)]
\end{equation}
\begin{equation}
\mathrm{IoU} = \frac{1}{|\mathcal{A}|}\sum_{a\in\mathcal{A}}
\frac{|\mathcal{T}_a \cap \hat{\mathcal{T}}_a|}
     {|\mathcal{T}_a \cup \hat{\mathcal{T}}_a|}
\end{equation}
We report the mean IoU across logs to account for variability in interaction
length.

\paragraph{Structural Validation.}
To evaluate whether the recovered attribution preserves transition structure,
we construct transition graphs and compare them via matrix similarity. Let $A^\star$ and $\hat{A}$ denote the ground-truth and predicted adjacency matrices, where $A^\star_{ij}\in\{0,1\}$ and $\hat{A}_{ij}=P_{ij}$. We measure structural similarity as:
\begin{equation}
\mathrm{EdgeSim} = \frac{\langle A^\star, \hat{A} \rangle}{\|A^\star\|_F \|\hat{A}\|_F}
\end{equation}

This metric computes the cosine similarity between ground-truth and predicted transition matrices, capturing both correct and spurious edges through aggregated edge-level confidences $P_{ij}$.

\paragraph{Robustness under Metadata Loss.}
To evaluate robustness under transcript corruption, we follow the evaluation protocol of the Who\&When benchmark~\citep{zhang2025whichagent}; additional benchmark details are provided in Appendix~\ref{app:who_when_dataset}.
Each failure trajectory provides annotations identifying the failure-responsible agent $a^\star$ and the decisive error step $t^\star$.  We simulate metadata-independent scenarios by applying two perturbations: 
(i) \textbf{ID Removal} and 
(ii) \textbf{Boundary Corruption}.
Given the corrupted transcript, our method first reconstructs agent identities and interaction boundaries, producing a recovered execution log. 
Failure attribution is then performed as a downstream task on the reconstructed structure using the Who\&When evaluation protocol. Given predictions $(\hat{a},\hat{t})$, we report:
\begin{equation}
\mathrm{AgentAcc} = \mathbb{I}[\hat{a}=a^\star], \qquad
\mathrm{StepAcc} = \mathbb{I}[\hat{t}=t^\star].
\end{equation}

AgentAcc measures whether the correct failure-responsible agent is identified, while StepAcc evaluates localization of the decisive error step. 
This setup treats failure attribution as a downstream task and measures whether reconstructed execution traces preserve the diagnostic utility of the original logs.

\begin{table*}[t]
\centering
\small
\setlength{\tabcolsep}{5pt}
\renewcommand{\arraystretch}{1.15}
\resizebox{\textwidth}{!}{
\begin{tabular}{l l | ccc | ccc | ccc}
\toprule
\multicolumn{2}{l|}{\cellcolor{headerblue}} &
\multicolumn{3}{c|}{\cellcolor{headerblue}\textbf{\#Agents = 4}} &
\multicolumn{3}{c|}{\cellcolor{headerblue}\textbf{\#Agents = 5}} &
\multicolumn{3}{c}{\cellcolor{headerblue}\textbf{\#Agents = 6}} \\
\multicolumn{2}{l|}{\cellcolor{headerblue}\textbf{Method \quad Metric}} &
\multicolumn{1}{c}{\cellcolor{headerblue}\textit{Star-Pure}} &
\multicolumn{1}{c}{\cellcolor{headerblue}\textit{Chain}} &
\multicolumn{1}{c|}{\cellcolor{headerblue}\textit{Tree}} &
\multicolumn{1}{c}{\cellcolor{headerblue}\textit{Star-Pure}} &
\multicolumn{1}{c}{\cellcolor{headerblue}\textit{Chain}} &
\multicolumn{1}{c|}{\cellcolor{headerblue}\textit{Tree}} &
\multicolumn{1}{c}{\cellcolor{headerblue}\textit{Star-Pure}} &
\multicolumn{1}{c}{\cellcolor{headerblue}\textit{Chain}} &
\multicolumn{1}{c}{\cellcolor{headerblue}\textit{Tree}} \\
\midrule

%--- End-to-end Attribution ---
\multicolumn{11}{l}{\cellcolor{secbanner}\textit{\textbf{End-to-end Attribution}}} \\
ChatGPT
  & IoU      & 0.191 & 0.185 & 0.178 & 0.157 & 0.143 & 0.133 & 0.111 & 0.110 & 0.119 \\
\cellcolor{white}
  & TokenAcc & 0.314 & 0.321 & 0.306 & 0.249 & 0.257 & 0.259 & 0.192 & 0.198 & 0.211 \\
\cellcolor{white}
  & EdgeSim  & \multicolumn{9}{c}{N/A} \\
\cellcolor{rowgray}DeepSeek
  & \cellcolor{rowgray}IoU      & \cellcolor{rowgray}0.175 & \cellcolor{rowgray}0.165 & \cellcolor{rowgray}0.178 & \cellcolor{rowgray}0.149 & \cellcolor{rowgray}0.140 & \cellcolor{rowgray}0.156 & \cellcolor{rowgray}0.111 & \cellcolor{rowgray}0.111 & \cellcolor{rowgray}0.125 \\
\cellcolor{rowgray}
  & \cellcolor{rowgray}TokenAcc & \cellcolor{rowgray}0.292 & \cellcolor{rowgray}0.279 & \cellcolor{rowgray}0.297 & \cellcolor{rowgray}0.251 & \cellcolor{rowgray}0.244 & \cellcolor{rowgray}0.263 & \cellcolor{rowgray}0.198 & \cellcolor{rowgray}0.198 & \cellcolor{rowgray}0.219 \\
\cellcolor{rowgray}
  & \cellcolor{rowgray}EdgeSim  & \multicolumn{9}{c}{\cellcolor{rowgray}N/A} \\
HMM / Viterbi
  & IoU
    & 0.614 & 0.609 & 0.607
    & 0.627 & 0.596 & 0.622
    & 0.635 & 0.579 & 0.637 \\
\cellcolor{white}
  & TokenAcc
    & 0.755 & 0.750 & 0.748
    & 0.762 & 0.742 & 0.757
    & 0.766 & 0.729 & 0.766 \\
\cellcolor{white}
  & EdgeSim
    & \underline{0.803} & \underline{0.796} & \underline{0.777}
    & \underline{0.820} & \underline{0.804} & \underline{0.802}
    & \underline{0.834} & \underline{0.811} & \underline{0.821} \\

\midrule

%--- Segmentation Methods ---
\multicolumn{11}{l}{\cellcolor{secbanner}\textit{\textbf{Segmentation Methods}}} \\
Recursive
  & IoU      & 0.556 & 0.451 & 0.529 & 0.489 & 0.385 & 0.483 & 0.430 & 0.306 & 0.458 \\
\cellcolor{white}
  & TokenAcc & 0.706 & 0.606 & 0.686 & 0.652 & 0.546 & 0.642 & 0.608 & 0.448 & 0.632 \\
\cellcolor{white}
  & EdgeSim  & \multicolumn{9}{c}{N/A} \\
\cellcolor{rowgray}Semantic-BoW
  & \cellcolor{rowgray}IoU      & \cellcolor{rowgray}0.305 & \cellcolor{rowgray}0.346 & \cellcolor{rowgray}0.329 & \cellcolor{rowgray}0.267 & \cellcolor{rowgray}0.315 & \cellcolor{rowgray}0.304 & \cellcolor{rowgray}0.247 & \cellcolor{rowgray}0.262 & \cellcolor{rowgray}0.281 \\
\cellcolor{rowgray}
  & \cellcolor{rowgray}TokenAcc & \cellcolor{rowgray}0.457 & \cellcolor{rowgray}0.517 & \cellcolor{rowgray}0.506 & \cellcolor{rowgray}0.442 & \cellcolor{rowgray}0.465 & \cellcolor{rowgray}0.469 & \cellcolor{rowgray}0.354 & \cellcolor{rowgray}0.380 & \cellcolor{rowgray}0.409 \\
\cellcolor{rowgray}
  & \cellcolor{rowgray}EdgeSim  & \multicolumn{9}{c}{\cellcolor{rowgray}N/A} \\
TextTiling
  & IoU      & 0.390 & 0.414 & 0.402 & 0.364 & 0.335 & 0.397 & 0.347 & 0.291 & 0.374 \\
\cellcolor{white}
  & TokenAcc & 0.513 & 0.568 & 0.540 & 0.474 & 0.498 & 0.471 & 0.428 & 0.406 & 0.468 \\
\cellcolor{white}
  & EdgeSim  & \multicolumn{9}{c}{N/A} \\

\midrule

%--- Signal-aware Watermarking (G3) ---
\multicolumn{11}{l}{\cellcolor{secbanner}\textit{\textbf{Signal-aware Watermarking}}} \\
Unigram
  & IoU
    & \underline{0.730} & \underline{0.714} & \underline{0.729}
    & 0.635 & 0.635 & \underline{0.660}
    & 0.615 & 0.612 & 0.634 \\
\cellcolor{white}
  & TokenAcc
    & \underline{0.833} & \underline{0.834} & \underline{0.842}
    & 0.764 & 0.766 & 0.783
    & 0.747 & 0.744 & 0.761 \\
\cellcolor{white}
  & EdgeSim
    & 0.570 & 0.590 & 0.530
    & 0.391 & 0.413 & 0.379
    & 0.336 & 0.351 & 0.332 \\
\cellcolor{rowgray}KGW
  & \cellcolor{rowgray}IoU
    & \cellcolor{rowgray}0.709
    & \cellcolor{rowgray}0.696
    & \cellcolor{rowgray}0.699
    & \cellcolor{rowgray}\underline{0.675}
    & \cellcolor{rowgray}\underline{0.652}
    & \cellcolor{rowgray}0.654
    & \cellcolor{rowgray}\underline{0.673}
    & \cellcolor{rowgray}\underline{0.656}
    & \cellcolor{rowgray}\underline{0.655} \\
\cellcolor{rowgray}
  & \cellcolor{rowgray}TokenAcc
    & \cellcolor{rowgray}0.827
    & \cellcolor{rowgray}0.816
    & \cellcolor{rowgray}0.819
    & \cellcolor{rowgray}\underline{0.804}
    & \cellcolor{rowgray}\underline{0.786}
    & \cellcolor{rowgray}\underline{0.787}
    & \cellcolor{rowgray}\underline{0.806}
    & \cellcolor{rowgray}\underline{0.791}
    & \cellcolor{rowgray}\underline{0.791} \\
\cellcolor{rowgray}
  & \cellcolor{rowgray}EdgeSim
    & \cellcolor{rowgray}\underline{0.603}
    & \cellcolor{rowgray}\underline{0.617}
    & \cellcolor{rowgray}\underline{0.578}
    & \cellcolor{rowgray}\underline{0.505}
    & \cellcolor{rowgray}\underline{0.488}
    & \cellcolor{rowgray}\underline{0.484}
    & \cellcolor{rowgray}\underline{0.461}
    & \cellcolor{rowgray}\underline{0.439}
    & \cellcolor{rowgray}\underline{0.399} \\
\midrule

%--- Our Method ---
\cellcolor{oursgreen}\textbf{IET}
  & \cellcolor{oursgreen}IoU $\uparrow$
    & \cellcolor{oursgreen}\textbf{0.814} & \cellcolor{oursgreen}\textbf{0.830} & \cellcolor{oursgreen}\textbf{0.834}
    & \cellcolor{oursgreen}\textbf{0.843} & \cellcolor{oursgreen}\textbf{0.839} & \cellcolor{oursgreen}\textbf{0.831}
    & \cellcolor{oursgreen}\textbf{0.819} & \cellcolor{oursgreen}\textbf{0.840} & \cellcolor{oursgreen}\textbf{0.835} \\
\cellcolor{oursgreen}
  & \cellcolor{oursgreen}TokenAcc $\uparrow$
    & \cellcolor{oursgreen}\textbf{0.866} & \cellcolor{oursgreen}\textbf{0.863} & \cellcolor{oursgreen}\textbf{0.867}
    & \cellcolor{oursgreen}\textbf{0.846} & \cellcolor{oursgreen}\textbf{0.851} & \cellcolor{oursgreen}\textbf{0.832}
    & \cellcolor{oursgreen}\textbf{0.830} & \cellcolor{oursgreen}\textbf{0.829} & \cellcolor{oursgreen}\textbf{0.818} \\
\cellcolor{oursgreen}
  & \cellcolor{oursgreen}EdgeSim $\uparrow$
    & \cellcolor{oursgreen}\textbf{0.929} & \cellcolor{oursgreen}\textbf{0.949} & \cellcolor{oursgreen}\textbf{0.935}
    & \cellcolor{oursgreen}\textbf{0.945} & \cellcolor{oursgreen}\textbf{0.942} & \cellcolor{oursgreen}\textbf{0.934}
    & \cellcolor{oursgreen}\textbf{0.912} & \cellcolor{oursgreen}\textbf{0.919} & \cellcolor{oursgreen}\textbf{0.921} \\
\bottomrule
\end{tabular}}
\caption{Sequential attribution performance across end-to-end attribution
baselines, segmentation methods, signal-aware watermarking baselines, and our method, under different agent counts and coordination
topologies. \textbf{Bold} highlights our method and
\underline{underline} the best-performing baseline for each metric and setting.}
\label{tab:exp1}
\end{table*}

\paragraph{Baselines.}
We study the problem of metadata-independent multi-agent attribution, where
only the final generated text is available and both agent identities and
execution logs are absent, and attribution is performed afterwards by an
offline auditor. To our knowledge, this setting has not been explicitly
formulated in prior work. Consequently, all baselines considered here are
adapted from related tasks (e.g., sequence labeling, segmentation, probabilistic
decoding, or text watermarking) and do not directly target this problem setting.
We compare against three categories of baselines.

\textbf{End-to-end attribution methods.}
For the LLM-based baselines, we use \textbf{gpt-5-mini} \citep{openai2025gpt5}
and \textbf{DeepSeek-v3.1} \citep{deepseekai2024deepseekv3}. Given the number of
agents $K$, we provide the model with reconstructed interaction text (assistant
outputs only, concatenated in a fixed order) and split it into fixed-length
token units. The LLM is prompted to assign token-span ranges to each speaker,
from which we derive token-level speaker labels. This setting requires the model
to jointly infer both segmentation boundaries and speaker ownership from text
alone. We also include \textbf{HMM / Viterbi}, a probabilistic baseline
where emission scores are defined by similarity to agent prototypes and a sticky
transition prior favors state persistence \citep{rabiner1989hmm,viterbi1967error}.

\textbf{Segmentation-based methods.}
We consider several segmentation-based methods in LangChain
\citep{langchain2022}: \textbf{Recursive}, which partitions text using simple
separators to match the target number of segments; \textbf{Semantic-BoW}, which
detects boundaries via drops in bag-of-words similarity between neighboring
windows; and \textbf{TextTiling}, a lexical-cohesion method based on local
similarity depth scores. These methods are evaluated only on segmentation
metrics (IoU and TokenAcc).

\textbf{Signal-aware watermarking methods.}
Unlike the semantic-only baselines, these methods embed a generation-time
signal that survives into the final text, making them the most directly
comparable provenance baselines for our setting. We assign each agent an
independent watermark key and recover attribution under the same pipeline and
metrics, adapting two representative schemes---%
\textbf{KGW}~\citep{kirchenbauer2023watermark,pan2024markllm} and
\textbf{Unigram}~\citep{Unigram}.

\begin{table*}[h]
\centering
\small
\setlength{\tabcolsep}{5pt}
\renewcommand{\arraystretch}{1.15}
\resizebox{\textwidth}{!}{
\begin{tabular}{l | cc cc cc | cc cc cc}
\toprule
% Row 1: top-level group headers
\cellcolor{headerblue} &
\multicolumn{6}{c|}{\cellcolor{headerblue}\textbf{Baseline{}}} &
\multicolumn{6}{c}{\cellcolor{oursgreen}\textbf{Ours}} \\
% Row 2: sub-group headers
\cellcolor{headerblue} &
\multicolumn{2}{c}{\cellcolor{headerblue}Clean} &
\multicolumn{2}{c}{\cellcolor{headerblue}Remove ID} &
\multicolumn{2}{c|}{\cellcolor{headerblue}Boundary} &
\multicolumn{2}{c}{\cellcolor{oursgreen}Clean} &
\multicolumn{2}{c}{\cellcolor{oursgreen}Remove ID} &
\multicolumn{2}{c}{\cellcolor{oursgreen}Boundary} \\
% Row 3: Method + Agent/Step headers
\cellcolor{headerblue}\textbf{Method} &
\cellcolor{headerblue}\textit{Agent} & \cellcolor{headerblue}\textit{Step} &
\cellcolor{headerblue}\textit{Agent} & \cellcolor{headerblue}\textit{Step} &
\cellcolor{headerblue}\textit{Agent} & \cellcolor{headerblue}\textit{Step} &
\cellcolor{oursgreen}\textit{Agent} & \cellcolor{oursgreen}\textit{Step} &
\cellcolor{oursgreen}\textit{Agent} & \cellcolor{oursgreen}\textit{Step} &
\cellcolor{oursgreen}\textit{Agent} & \cellcolor{oursgreen}\textit{Step} \\
\midrule
% Row: All-at-Once
All-at-Once
  & 54.33 & 12.50
  & 4.17  & 9.69
  & 24.54 & 9.19
  & \cellcolor{white}54.16 & \cellcolor{white}10.47
  & \cellcolor{white}24.45 & \cellcolor{white}10.71
  & \cellcolor{white}36.67 & \cellcolor{white}9.54 \\
% Row: Step-by-Step (gray alternating)
\cellcolor{rowgray}Step-by-Step
  & \cellcolor{rowgray}35.20 & \cellcolor{rowgray}25.51
  & \cellcolor{rowgray}0.00  & \cellcolor{rowgray}10.28
  & \cellcolor{rowgray}33.54 & \cellcolor{rowgray}14.41
  & \cellcolor{rowgray}33.13 & \cellcolor{rowgray}23.51
  & \cellcolor{rowgray}21.51 & \cellcolor{rowgray}14.71
  & \cellcolor{rowgray}37.00 & \cellcolor{rowgray}18.84 \\
% Row: Binary Search
Binary Search
  & 44.13 & 23.98
  & 0.00  & 3.17
  & 31.71 & 1.59
  & \cellcolor{white}42.71 & \cellcolor{white}21.00
  & \cellcolor{white}21.98 & \cellcolor{white}4.45
  & \cellcolor{white}35.28 & \cellcolor{white}2.38 \\
\bottomrule
\end{tabular}
}
\caption{Failure attribution accuracy under metadata corruption.
Left: baseline methods without trace signals.
Right: our tracing-enabled method.
\textit{Agent} and \textit{Step} denote agent-level and step-level attribution accuracy (\%).}
\label{tab:exp2}
\end{table*}

\section{Result}

Our evaluation asks four questions. The first concern is whether embedding
provenance signals at generation time degrades the reasoning performance that
multi-agent inference is deployed to obtain. The remaining three concerns are what
the audit setting requires---recovering token-level attribution, reconstructing
the transition structure induced by the output, and preserving downstream
diagnostic utility once execution metadata is gone.

\begin{figure}[t]
  \centering
  \includegraphics[width=\linewidth]{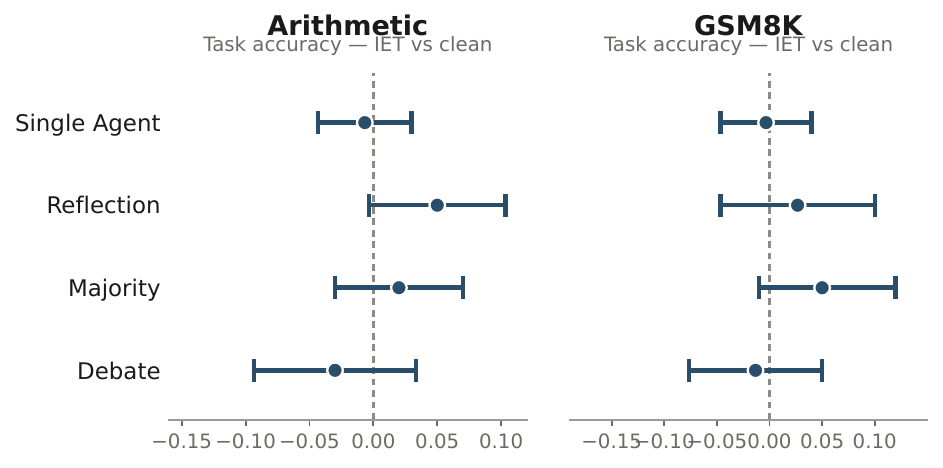}
  \caption{\textbf{IET preserves task performance}}
  \label{fig:quality-impact}
\end{figure}

\paragraph{Will IET degrade general performance?}
To assess whether watermarking degrades the reasoning gains of multi-agent inference, we
adopt the same four inference-time strategies evaluated in Table~1 of
\citet{du2023improving}: \textit{Single Agent}, \textit{Single Agent (Reflection)},
\textit{Multi-Agent (Majority)}, and \textit{Multi-Agent (Debate)}. We follow their strategy
definitions but substitute \texttt{Llama-3.1-8B-Instruct} as the base model, running debate
with 3 agents for 2 rounds. We evaluate each strategy on the same two reasoning benchmarks,
Arithmetic and Grade School Math (GSM8K), under clean generation and under our IET. For each
task--method cell we report the difference together with its 95\% confidence interval. As
shown in Figure~\ref{fig:quality-impact}, every difference is small and its confidence interval
overlaps zero, indicating that IET does not significantly change task accuracy for any
strategy or benchmark.

\paragraph{How accurately can sequential attribution be recovered?}
We report TokenAcc and IoU across different agent counts and interaction
topologies in Table~\ref{tab:exp1}. LLM-based baselines (ChatGPT and DeepSeek)
perform poorly, with TokenAcc below $0.33$ and IoU in the $0.11$--$0.19$ range,
and both degrade further as the agent pool grows. Segmentation methods, which
locate boundaries from lexical cohesion alone, improve on this but remain far
from reliable attribution: Recursive peaks at IoU $0.556$ on star-pure with
four agents and falls to $0.306$ on chain with six. HMM/Viterbi is the
strongest semantic-only baseline (IoU $0.579$--$0.637$, TokenAcc
$0.729$--$0.766$), and the signal-aware baselines are stronger still
(IoU $0.612$--$0.730$, TokenAcc $0.744$--$0.833$). IET attains IoU
$0.814$--$0.843$ and TokenAcc $0.818$--$0.867$ across all nine settings.

\paragraph{How well does the recovered trace preserve transition structure?}
Table~\ref{tab:exp1} also reports transition structure accuracy (EdgeSim). Our
method achieves consistently high accuracy across all configurations
($0.912$--$0.949$) and stays well above the strongest baseline at every agent
count (margins of $0.078$--$0.158$). Differences across topologies within a
given agent count are small ($0.002$--$0.020$ pairwise) and follow no
consistent ordering, indicating that the linearized transition structure is
recovered comparably well under chain, star, and tree coordination.

\paragraph{Can the recovered trace support downstream failure attribution under
metadata loss?}
Table~\ref{tab:exp2} shows that IET's provenance signal is \emph{nearly free on
clean transcripts yet decisive once metadata is lost}. On the clean columns,
where no reconstruction is involved, instrumentation costs at most $2.1$
AgentAcc and $3.0$ StepAcc points. The payoff is stark under corruption. When agent identities are
stripped, the transcript names no one and the baselines collapse: AgentAcc
falls to at most $4.17$, and to exactly $0.00$ for two of the three protocols,
making attribution impossible. IET instead recovers the missing identities and
restores AgentAcc to $21.5$--$24.5$, retaining $45$--$65\%$ of its own clean
ceiling and turning a near-zero baseline into usable attribution. Under the
milder boundary corruption IET again dominates, outperforming the baseline on
\emph{every} protocol and \emph{both} metrics (AgentAcc by up to $+12.1$
points) and roughly halving the attribution loss wherever the baseline
degrades. The trend is unambiguous: the more metadata a corruption destroys,
the wider IET's margin.

\vspace{-5pt}
\section{Conclusion}
We introduced IET, a provenance-by-design
framework for auditing multi-agent systems under metadata loss. By
embedding signals directly into the token distribution at
generation time, IET turns the output into a self-verifying record,
enabling an auditor holding only a verification registry to recover
segment boundaries, per-segment agent attribution, and the induced transition
structure from the final text alone. Our experiments show that provenance recovered this way remains accurate across
coordination topologies and agent counts, survives privacy-preserving
redaction, and restores substantial downstream failure-attribution capability. As agent-generated text routinely outlives the environment that produced it,
carrying provenance inside the artifact offers a practical foundation for
accountability under metadata loss.

% *ACL requires an unnumbered Limitations section (does not count
% toward the page limit).
\newpage
\section*{Limitations}
\label{sec:limitations}

IET is a provenance-by-design framework: it assumes participating agents apply keyed decoding-time instrumentation and that an offline auditor holds the corresponding verification registry, so it is not a universal post-hoc attribution method for arbitrary text. The signal is carried at the token level and is designed to survive metadata loss, PII redaction, and moderate text transformation; heavier semantic rewriting progressively erodes it, and adversarial suppression or forgery of provenance signals is outside our current scope. 

\section*{Broader Impact}

IET introduces a way to audit interactions among agents that may be
operated by different companies or parties, establishing
accountability from the final text alone without requiring anyone to
expose private logs, weights, or execution traces. By binding each
agent's identity to its output at generation time, it offers a
privacy-preserving foundation for oversight when interaction metadata
is unavailable, and points toward future work on cross-organizational
provenance, verifiable inter-agent contracts, and accountability
standards for open multi-agent ecosystems. Because IET assumes
cooperative instrumentation and does not target adversarial agents,
we see no significant risk of direct negative societal impact from
this method.

\section*{Ethical Considerations}

Our research focuses on attribution and execution tracing in multi-agent language systems. To systematically evaluate our approach, we conduct experiments on conversational datasets that may contain personally identifiable information (PII), such as names or contact details. These datasets are obtained from publicly available resources and are used solely for research purposes in a controlled experimental setting. To mitigate privacy risks, we apply standard PII redaction procedures by replacing sensitive spans with placeholder tokens (e.g., [NAME], [EMAIL]). Our work aims to improve transparency and accountability in multi-agent AI systems by enabling reliable attribution of generated content, which aligns with responsible AI development and auditing practices.

% -------------------------------------------------------------
%  Bibliography: the ACL template uses acl_natbib.bst via a plain
%  \bibliography call (no \bibliographystyle needed).
%  Rename your .bib file to custom.bib, or change the name below
%  to match your existing colm2026_conference.bib.
% -------------------------------------------------------------
\bibliography{custom}

\appendix

\label{sec:appendix}

% =========================================================
% Appendix
% =========================================================
\appendix

\section{Related Work}

\subsection{Failure Attribution in Multi-Agent Systems}

As large language models are increasingly deployed in multi-agent settings, understanding and debugging system failures has become an important research problem \citep{chen2024agentverse,wang2026trajectorysurvey,huang2025deepresearchguard,zhou-etal-2024-large-language}.
Prior work primarily studies \emph{failure attribution} as a post-hoc inference task: given an execution trace produced by multiple interacting agents, the goal is to infer which agent, or which step in the interaction, is responsible for a system-level failure \citep{zhang2025whichagent,ge2025failureattributing,barke2026agentrx,cemri2025multifail}.
Recent benchmarks formalize this problem explicitly by asking models to predict the responsible agent (\emph{who}) and the decisive error step (\emph{when}) from raw execution traces.
Other approaches employ counterfactual replay, causal analysis \cite{zhang2025agentracer}, or learned tracer models \cite{zhu2025raffles,zhang2025graphtracer} to identify failure sources in the absence of explicit provenance.
These methods highlight the difficulty of attribution when execution
logs record only surface-level interactions without preserving information-flow dependencies.
 Our approach is complementary: instead of relying solely on retrospective inference, we embed traceability into the generation process so that agent-level provenance can be recovered more directly.
\vspace{-8pt}
\subsection{Watermarking and Provenance in Large Language Models}

Watermarking has been widely studied for identifying AI-generated
content and verifying model provenance \cite{fang2025muse}. Existing methods embed
detectable signals at different stages of the generation pipeline,
including during logits generation
\citep{kirchenbauer2023watermark,lee2024codewatermarking,hu2023unbiasedwatermark,wu2024dipmark},
token sampling
\citep{christ2024undetectable,kuditipudi2024distortionfree,hou2024semstamp},
or model training
\citep{sun2022coprotector,sun2023codemark,gu2023learnabilitywatermarks}.
These approaches primarily focus on detecting whether a piece of text
was generated by a particular model, and are typically designed for
single-model settings rather than tracing information flow across
multiple agents. 

A separate line of work studies watermarking for existing text through
format-based, lexical-based, syntactic-based, and generation-based
transformations
\citep{brassil1995electronicmarking,topkara2006ambiguitywatermarking,atallah2001nlwatermarking,
abdelnabi2021adversarialwatermarking}. However, these approaches embed signals into a single text instance and do not model how provenance propagates across multiple interacting agents or intermediate steps. In contrast, IET is
a \emph{system-level instrumentation mechanism} that preserves provenance across agent interactions, enabling explicit and
verifiable attribution in multi-agent systems.

\section{Prompt Details}
\label{app:prompt_details}

This section presents the full prompt templates used in our experiments.
We organize them into two groups:
(1) prompts used in the LLM baseline for speaker-range assignment, and
(2) prompts used in the debate framework, including the debating agents and the final judge.
Placeholders enclosed in braces (e.g., \{question\_text\}, \{choices\_block\}) denote instance-specific fields filled dynamically at inference time.

\begin{table*}[t]
\centering
\small
\setlength{\tabcolsep}{4pt}
\renewcommand{\arraystretch}{1.15}
\begin{tabular}{%
  >{\raggedright\arraybackslash}p{0.12\textwidth}
  >{\raggedright\arraybackslash}p{0.19\textwidth}
  >{\raggedright\arraybackslash}p{0.30\textwidth}
  >{\raggedright\arraybackslash}p{0.27\textwidth}}
\toprule
\cellcolor{headerblue}\textbf{Dataset} &
\cellcolor{headerblue}\textbf{Subset / File} &
\cellcolor{headerblue}\textbf{Source} &
\cellcolor{headerblue}\textbf{Role in Experiments} \\
\midrule
Arithmetic
  & synthetic
  & \seqsplit{composable-models/llm\_multiagent\_debate} (GH)
  & Generation quality evaluation \\
\cellcolor{rowgray}GSM8K
  & \cellcolor{rowgray}\texttt{test}
  & \cellcolor{rowgray}\seqsplit{openai/gsm8k} (HF), via \seqsplit{composable-models/llm\_multiagent\_debate} (GH)
  & \cellcolor{rowgray}Generation quality evaluation \\
MAMA topology
  & \seqsplit{llama3.1\_num484\_nopii.csv}
  & \citet{liu2025topologymatters}
  & Topology-based multi-agent experiments \\
\cellcolor{rowgray}Who \& When
  & \cellcolor{rowgray}\texttt{Algorithm-Generated}, \texttt{Hand-Crafted}
  & \cellcolor{rowgray}\seqsplit{mingyin1/Agents\_Failure\_Attribution} (GH)
  & \cellcolor{rowgray}Failure attribution under metadata corruption \\
\bottomrule
\end{tabular}
\caption{Datasets used in different stages of our experiments. HF: HuggingFace Hub; GH: GitHub.}
\label{tab:dataset_summary}
\end{table*}

% =========================================================
% LLM Baseline Prompts
% =========================================================

\subsection{LLM Baseline Prompts}
\label{app:llm_baseline_prompts}
\label{app:baseline_overview}

This section presents the prompt templates used in the LLM baseline for
speaker-range assignment. Placeholders enclosed in braces (e.g.,
\{unit\_tokens\}) denote instance-specific fields filled dynamically at
inference time.

The baseline prompt consists of two parts: a system prompt and a user prompt.
The system prompt specifies the output schema and the global validity constraints, including complete coverage, no overlap, boundary alignment, and non-empty assignment for every speaker.
The user prompt provides the instance-specific information, including the number of speakers, the number of units, the total number of tokens, the valid unit boundaries, the task description, and the unit-level trace content.

Together, these prompts cast the attribution task as a constrained structured prediction problem: the model must first assign each unit to exactly one speaker and then merge adjacent units belonging to the same speaker into minimal non-overlapping ranges.
The full prompt templates are provided below.

\paragraph{LLM Baseline Prompt (System Prompt).}
\label{app:baseline_system_prompt}
\noindent\textit{System prompt template used in the LLM baseline for speaker-range assignment.}

\begin{quote}
\begin{Verbatim}[fontsize=\scriptsize,breaklines=true,breakanywhere=true]
You assign each token span in a concatenated multi-speaker trace to one of K speakers.
You must reason privately and NEVER reveal your reasoning.
Your final answer must be EXACTLY ONE JSON object and NOTHING ELSE.
Do not output any natural language, analysis, prefacing text, markdown, code fences, bullets, or apologies.
Do not start with phrases like 'Okay', 'Let's', 'Here is', or any sentence.
The ONLY valid output format is:
{"ranges": {"0": [[0, 64]], "1": [[64, 128]], "2": [[128, 192]], "3": [[192, 256]]}}
Example valid output:
{"ranges": {"0": [[0, 64]], "1": [[64, 128], [192, 256]], "2": [[128, 192]], "3": [[256, 320]]}}
Use speaker ids 0..K-1 in order of first appearance in the conversation.
Each [l, r] is a token range in [start, end) format.
Each listed unit is a contiguous chunk of {unit_tokens} tokens, except the final chunk of a message which may be shorter.
Every range boundary must align exactly to a listed unit boundary.
That means every start and every end value must be chosen from the provided unit boundary values.
Do not invent token boundaries inside a unit.
Think of the task as assigning each unit to exactly one speaker, then merging consecutive units with the same speaker.
No unit may belong to more than one speaker.
No unit may be left unassigned.
The top-level object must contain exactly one key: "ranges".
Inside "ranges", there must be exactly K keys: "0", "1", ..., "K-1".
No speaker key may appear outside the "ranges" object.
Ranges for each speaker must be sorted and non-overlapping.
For each speaker, merge all adjacent or touching ranges.
If one range ends at x and the next range starts at x, they must be merged into a single range.
Use the minimum number of ranges possible for each speaker.
Across all speakers, ranges must exactly cover all tokens from 0 to T with no gaps and no overlaps.
Every speaker from 0 to K-1 definitely appears in this trace.
Therefore, no speaker may have an empty range list.
Any output where a speaker has [] is invalid.
Any output where one speaker covers all tokens is invalid.
Before answering, verify that every speaker has at least one non-empty range with positive length.
If you are uncertain, still output your best guess in the required JSON format.
An answer like 'Okay, let me think' is invalid.
Any text before '{' or after '}' is invalid.
\end{Verbatim}
\end{quote}

\paragraph{LLM Baseline Prompt (User Prompt).}
\label{app:baseline_user_prompt}
\noindent\textit{User prompt template used in the LLM baseline for speaker-range assignment. Instance-specific fields are filled dynamically at inference time.}

\begin{quote}
\begin{Verbatim}[fontsize=\scriptsize,breaklines=true,breakanywhere=true]
There are K={k} speakers.
There are N={num_units} units, indexed from 0 to {num_units - 1}.
There are T={total_tokens} tokens in total.
Each unit is a contiguous chunk of {unit_tokens} tokens, except the last chunk of a message which may be shorter.
Valid unit boundary values are: {boundary_values}.
You do NOT know the speaker identities in advance.
Speaker 0 must be the first distinct speaker that appears, speaker 1 the next new speaker, and so on.
Return exactly one JSON object with this schema:
{"ranges": {"0": [[start, end]], "1": [[start, end]], "2": [[start, end]], "3": [[start, end]]}}
Concrete example output:
{"ranges": {"0": [[0, 64]], "1": [[64, 128], [192, 256]], "2": [[128, 192]], "3": [[256, 320]]}}
In that example, speaker 1 has two ranges: [64, 128] and [192, 256].
All start/end values in the output must be taken from the valid unit boundary values listed above.
A range like [10, 70] is invalid if 10 or 70 is not a listed unit boundary.
Assign each unit to exactly one speaker before you construct the ranges.
Then merge adjacent units assigned to the same speaker into a single range.
There are exactly K distinct speakers in this trace.
You must include every speaker key from 0 to K-1, and every speaker must have at least one non-empty range.
Every speaker definitely speaks at least once in this trace.
So no speaker may have an empty list [].
The JSON is invalid unless the only top-level key is "ranges" and all speaker keys are inside it.
The JSON is invalid if any speaker key from 0 to K-1 is missing.
The JSON is invalid if any speaker has an empty list.
The JSON is invalid if one speaker covers all tokens.
Do not assign all tokens to a single speaker.
Adjacent or touching ranges for the same speaker must be merged.
Use the minimum number of ranges possible.
Your first character must be '{' and your last character must be '}'.
Do not output anything except this JSON object.

TRACE CONTEXT:
task_description={task_description}

UNITS:
{unit_lines}
\end{Verbatim}
\end{quote}

\section{Dataset Details}
\label{app:dataset_details}

This section summarizes the datasets used in different stages of our experiments.

% \subsection{C4 Dataset for Initial Watermark Testing}
% \label{app:c4_dataset}

% For the initial evaluation of the Waterfall watermark, we used the \texttt{realnewslike} configuration of the C4 dataset (\texttt{allenai/c4}) \citep{raffel2020t5}.
% This dataset was used as a source of natural-language text for preliminary watermarking experiments before moving to the main multi-agent trace setting.

% \subsection{TruthfulQA for Debate Evaluation}
% \label{app:truthfulqa_dataset}

% For the debate experiments, we used the multiple-choice setting of TruthfulQA \citep{lin2022truthfulqa}.
% We selected this dataset because it provides factual questions with multiple candidate answers, which aligns naturally with our debate framework.
% In each instance, two agents debate which option is the most truthful and least misleading, and a judge model then makes the final selection based on the full debate transcript.

% \subsection{HotpotQA for Sliding-Window and Concatenation Experiments}
% \label{app:hotpotqa_dataset}

% For the sliding-window experiments and the concatenated-text watermark detection experiments, we used the \texttt{distractor} configuration of the HotpotQA dataset (\texttt{hotpot\_qa}) with the \texttt{train} split \citep{yang2018hotpotqa}.
% This dataset served as a source of natural-language passages for constructing the inputs used in these experiments.
% We used samples from this dataset to build longer composite inputs by concatenating text segments, allowing us to evaluate watermark detectability under boundary shifts and mixed-context settings.

\subsection{MAMA Topology Dataset}
\label{app:mama_topology_dataset}

For the topology experiments, we used the dataset file \texttt{llama3.1\_num484\_nopii.csv}, which was provided by the MAMA repository owner.
We inspected this file together with the public description of the MAMA project to understand its role in the experimental pipeline \citep{liu2025topologymatters}.

MAMA is designed to measure privacy leakage in multi-agent LLM systems as a function of communication topology.
According to its public description, the framework starts from synthetic documents containing labeled PII entities, generates sanitized task instructions, and then evaluates leakage under different graph structures such as fully connected, ring, chain, binary tree, star, and star-ring topologies \citep{liu2025topologymatters}.

\begin{table}[t]
\centering
\small
\setlength{\tabcolsep}{5pt}
\renewcommand{\arraystretch}{1.15}
\begin{tabular}{p{0.34\columnwidth} p{0.50\columnwidth}}
\toprule
\cellcolor{headerblue}\textbf{Attribute} &
\cellcolor{headerblue}\textbf{Value} \\
\midrule
File name
  & \texttt{llama3.1\_num484\_nopii.csv} \\
\cellcolor{rowgray}Number of examples
  & \cellcolor{rowgray}484 \\
Columns
  & \texttt{text}, \texttt{pii}, \texttt{generated\_texts}, \texttt{task\_backgrounds}, \texttt{questions} \\
\cellcolor{rowgray}Data granularity
  & \cellcolor{rowgray}Instance-level source records \\
Intended role
  & Source material for topology-based multi-agent experiments \\
\bottomrule
\end{tabular}
\caption{Basic statistics of the MAMA topology dataset file used in our experiments.}
\label{tab:mama_dataset_stats}
\end{table}

\subsection{Who \& When Benchmark}
\label{app:who_when_dataset}

For robustness evaluation under metadata loss, we use the \textit{Who \& When} benchmark released in the Agents Failure Attribution repository \citep{zhang2025whichagent}.
According to the repository documentation, the benchmark contains 184 annotated failure tasks collected from two sources:
(i) algorithm-generated multi-agent systems built using CaptainAgent, and
(ii) hand-crafted systems such as Magnetic-One.
Each failure case is annotated with the failure-responsible agent, the decisive error step, and a natural-language explanation of the failure.

In the repository, the dataset is organized under the \texttt{Who\&When} directory, which contains two subdirectories:
\texttt{Algorithm-Generated} and \texttt{Hand-Crafted}.
Each sample is stored as a separate JSON file.
Inspection of representative examples shows that the records contain a conversation or execution \texttt{history} together with task-level supervision fields such as \texttt{question}, \texttt{ground\_truth}, \texttt{question\_ID}, \texttt{mistake\_agent}, \texttt{mistake\_step}, and \texttt{mistake\_reason}.
We also observe minor schema variation across subsets: for example, the algorithm-generated subset includes fields such as \texttt{is\_correct} and \texttt{level}, whereas the hand-crafted subset uses \texttt{history} as the main trajectory container and may differ slightly in auxiliary field naming.

For our experiments, we use this benchmark as a failure-attribution testbed under transcript corruption.
The original annotations provide the ground-truth responsible agent and decisive error step, while the recorded \texttt{history} field supplies the multi-agent trajectory on which restoration and downstream attribution are evaluated.

\begin{table}[t]
\centering
\small
\setlength{\tabcolsep}{5pt}
\renewcommand{\arraystretch}{1.15}
\begin{tabular}{p{0.30\columnwidth} p{0.62\columnwidth}}
\toprule
\cellcolor{headerblue}\textbf{Attribute} &
\cellcolor{headerblue}\textbf{Value} \\
\midrule
Benchmark name
  & \textit{Who \& When} \\
\cellcolor{rowgray}Repository organization
  & \cellcolor{rowgray}\begin{tabular}[t]{@{}l@{}}
      \texttt{Who\&When/Algorithm-Generated} \\
      \texttt{Who\&When/Hand-Crafted}
    \end{tabular} \\
Number of failure tasks
  & 184 annotated failure tasks \\
\cellcolor{rowgray}Storage format
  & \cellcolor{rowgray}One JSON file per task instance \\
Core supervision
  & Failure-responsible agent, decisive error step, natural-language explanation \\
\cellcolor{rowgray}Representative fields
  & \cellcolor{rowgray}\texttt{history}, \texttt{question}, \texttt{ground\_truth}, \texttt{question\_ID}, \texttt{mistake\_agent}, \texttt{mistake\_step}, \texttt{mistake\_reason} \\
Intended role
  & Failure attribution under metadata corruption and transcript restoration \\
\bottomrule
\end{tabular}
\caption{Summary of the \textit{Who \& When} benchmark used for robustness evaluation.}
\label{tab:who_when_dataset_stats}
\end{table}

In summary, the Who \& When benchmark provides annotated multi-agent failure trajectories with ground-truth labels for both the responsible agent and the decisive error step, making it suitable for evaluating whether recovered traces preserve the downstream diagnostic utility of the original execution logs.

\subsection{Reasoning Benchmarks for Generation Quality}
Arithmetic follows \citet{du2023improving}: we sample 100 expressions of the
form $a+b\cdot c+d-e\cdot f$ with $a,\dots,f$ drawn uniformly from
$\{0,\dots,30\}$ (seed 42). For GSM8K we use 100 problems from the test split.

\section{Experimental Details}
\label{app:exp_details}

This appendix provides additional implementation details for the experiments reported in the main paper, including dataset setup, generation and detection parameters, corruption protocols, and supplementary evaluation settings.

\subsection{MAMA Topology Tracing Experiments}
\label{app:mama_exp_details}

Our main topology-controlled tracing experiments are conducted on the Multi-Agent Interaction Dataset using the file
\texttt{dataset/llama3.1\_num484\_nopii.csv}.
We evaluate three communication topologies:
\texttt{chain}, \texttt{star\_pure}, and \texttt{tree},
under three agent-count settings: 4, 5, and 6 agents.
Each condition contains 484 samples.

For all MAMA experiments, we use
\texttt{meta-llama/Llama-3.1-8B-Instruct}
as the actual generation model.
In the experiment scripts, the target node is fixed to
\texttt{target\_idx = 0},
the attacker node is fixed to
\texttt{attacker\_idx = 3},
and the maximum number of rounds is
\texttt{max\_rounds = 3}.

\paragraph{Watermarking configuration.}
We use the Fourier watermark function with
\texttt{k\_p = 1}
and
\texttt{kappa = 0.8}.
Agent-specific watermark IDs are assigned consecutively starting from 42.
Unless otherwise stated, the watermark detector uses
\texttt{n\_gram = 2}.

\paragraph{Detection and boundary reconstruction.}
Unless otherwise noted, the sliding-window detector uses
\texttt{window\_tokens = 64} and \texttt{step\_tokens = 16}. The smoothing and
local boundary-selection parameters are \texttt{smooth\_win = 5},
\texttt{local\_radius = 8}, and \texttt{min\_points\_for\_pair = 10}. For
visualization and score summarization, we use \texttt{hist\_bins = 30} and
\texttt{worst\_seq\_k = 5}.

\paragraph{Transition structure analysis.}
Recovered traces are used to extract transition patterns between agents and summarize them as a transition graph.
The implementation supports classification into several canonical coordination patterns, including
\texttt{complete},
\texttt{tree},
\texttt{chain},
\texttt{star\_pure},
\texttt{star\_ring},
and
\texttt{circle}.
The main paper reports results for
\texttt{chain},
\texttt{star\_pure},
and
\texttt{tree}.
The definition of transition structure accuracy (EdgeSim) follows the formulation introduced in the main text.

\subsection{Baseline Details for MAMA}
\label{app:mama_baseline_details}

We compare our method against both LLM-based attribution baselines and traditional segmentation baselines.

\paragraph{LLM baselines.}
The codebase supports multiple API-based LLM baselines, including OpenAI, DeepSeek, Qwen, and Bedrock backends.
For the reported runs, the common baseline configuration uses
\texttt{unit\_mode = assistant\_response},
\texttt{unit\_tokens = 64},
\texttt{temperature = 0.0},
and
\texttt{top\_p = 1.0}.
This setup asks the model to assign token-span ownership to one of the candidate agents based only on the final concatenated interaction text.

\paragraph{Segmentation baselines.}
We include three segmentation baselines:
\textbf{Recursive},
\textbf{Semantic-BoW},
and
\textbf{TextTiling}.
The default baseline configuration uses
\texttt{unit\_mode = assistant\_response},
\texttt{seed = 123},
and
\texttt{max\_files = 0}
(process all available files).

\subsection{Failure Attribution Under Metadata Loss on Who\&When}
\label{app:who_when_exp_details}

For failure attribution under metadata loss, we use the Who\&When benchmark.
The benchmark contains 184 annotated failure trajectories in total, including
126 \textbf{Algorithm-Generated} trajectories and
58 \textbf{Hand-Crafted} trajectories.
The local evaluation pipeline primarily targets the 126 algorithm-generated trajectories, which are also the default input subset in the watermark-based scripts. 

\paragraph{Downstream attribution methods.}
We evaluate three downstream failure-attribution protocols:
\texttt{all\_at\_once},
\texttt{step\_by\_step},
and
\texttt{binary\_search}.
For the reported results in Table~\ref{tab:exp2}, the evaluator model is
\texttt{meta-llama/Llama-3.1-8B-Instruct}.
The evaluation metrics are
\textbf{AgentAcc} and
\textbf{StepAcc}, following the Who\&When protocol.

\paragraph{ID removal protocol.}
To simulate metadata-independent conditions, we remove agent-specific identity signals from the trajectories.
For algorithm-generated traces, human steps are identified by missing or empty \texttt{name} fields, while non-human steps are identified by the presence of an agent name.
After anonymization, human turns are normalized to a generic human identity, while non-human agent turns are collapsed into a shared generic assistant identity.
Concretely, human turns are assigned
\texttt{role = "human"}
with an empty name field, and non-human turns are assigned
\texttt{role = "assistant"}
and
\texttt{name = "Agent"}.
When present, top-level identity-leaking fields such as \texttt{system\_prompt} are removed.

For hand-crafted traces, human turns are identified by
\texttt{role == "human"}.
After anonymization, human turns remain generic human turns, while all non-human turns are mapped to a generic assistant role and their \texttt{name} fields are removed.

\paragraph{Boundary corruption protocol.}
For boundary corruption, we preserve the number of turns, their order, and their associated metadata, but corrupt the content boundaries.
Specifically, all turn contents in a trajectory are concatenated into a single text stream using \texttt{\textbackslash n\textbackslash n} as the separator.
The resulting stream is then randomly re-segmented into the original number of turns using random cut points, and the new segments are refilled back into the original turn skeleton.
This random cut-and-refill procedure keeps the trace skeleton unchanged while corrupting turn boundaries.
We use random seed 42 for this re-segmentation protocol.

\section{Case Study: Robustness to PII Redaction}
\label{sec:pii-redaction}

We further examine whether privacy-preserving preprocessing affects
trace recovery. In the Multi-Agent Interaction Dataset, conversation
logs may contain sensitive information such as names, emails, or phone
numbers. To simulate realistic deployment settings, we apply standard
PII redaction that replaces sensitive spans with placeholder tokens
(e.g., [NAME], [EMAIL]).

\begin{figure}[t]
  \centering
  \includegraphics[width=\linewidth]{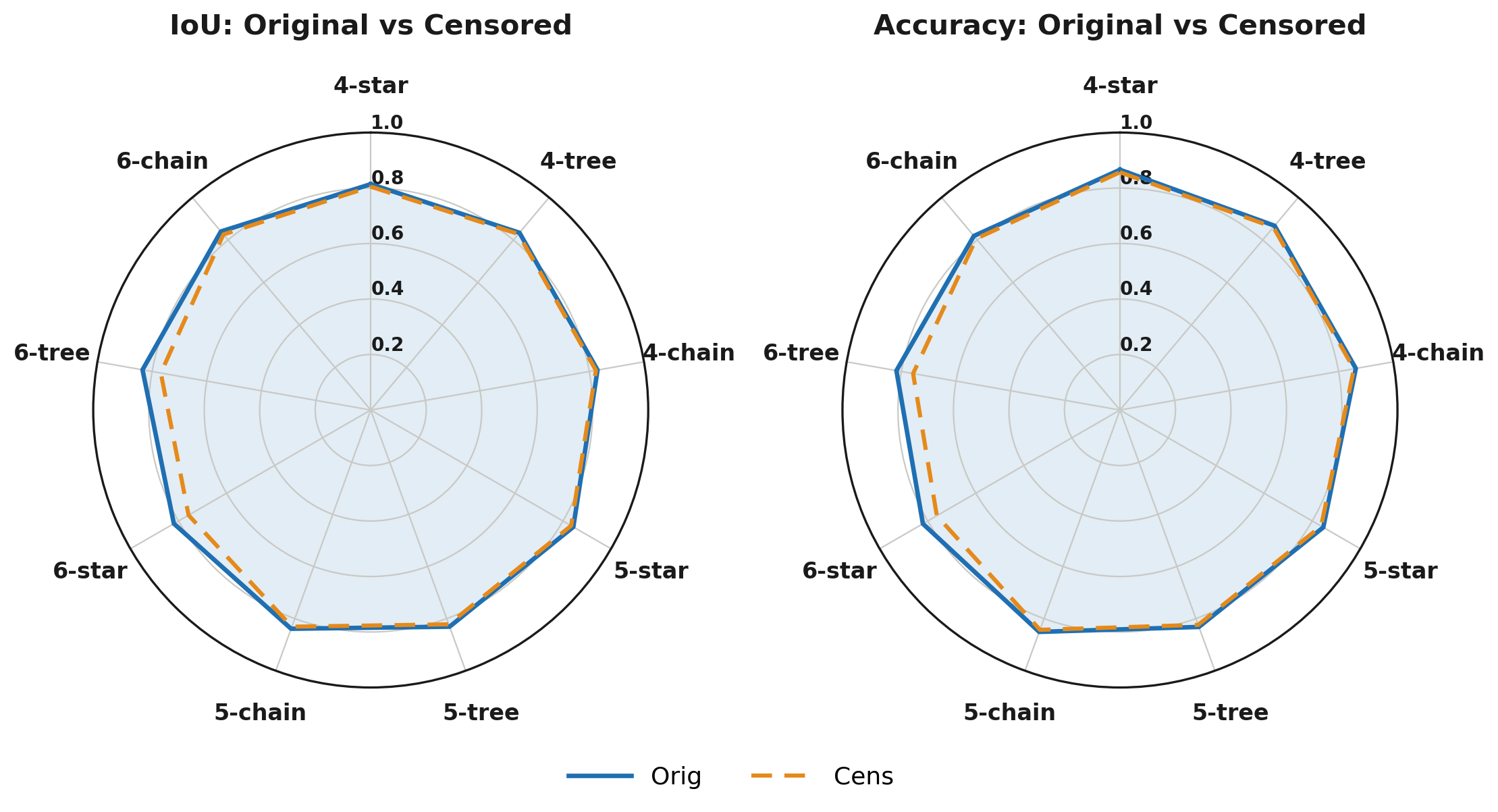}
  \caption{\textbf{Original vs.\ PII-redacted attribution performance.}
  IoU (left) and token-level accuracy (right) of IET across nine
  topology--agent-count settings (Star-Pure, Chain, Tree $\times$ 4, 5,
  6 agents), computed on the original transcripts (\textit{Orig}) and
  on the same transcripts after PII redaction (\textit{Cens}). The two
  curves nearly overlap everywhere except 6-star-pure and 6-tree,
  where redaction causes a small, localized drop.}
  \label{fig:pii-redaction}
\end{figure}

Figure~\ref{fig:pii-redaction} compares tracing performance before and
after redaction across different multi-agent interaction structures.
Despite removing lexical content, overall performance remains nearly
unchanged: across all nine topology--agent-count settings, IoU and
token-level accuracy stay within a narrow band under both conditions.
For most interaction patterns (e.g., 4-chain, 4-tree, and 5-chain), the
IoU and token-level accuracy curves for the original and censored logs
almost overlap. The only noticeable deviations appear in the more
complex 6-star-pure and 6-tree settings, where redaction produces a
small but visible drop in both metrics. These results indicate that
the tracing signal is embedded in the generation process itself
rather than in surface lexical patterns. Consequently, removing
sensitive tokens does not substantially affect agent attribution,
demonstrating that the proposed tracing mechanism is robust to
privacy-preserving transcript redaction.

\section{Sensitivity to Post-Generation Text Transformations}
\label{sec:transform-sensitivity}

Beyond metadata-level corruption (ID removal, boundary corruption, PII
redaction), agent-generated text may also be altered after generation
through ordinary processing. We characterize IET's sensitivity to
such transformations using the Oracle-IET protocol (ground-truth
boundaries with keyed attribution scoring) on \texttt{chain\_4}.
Table~\ref{tab:transform_sensitivity} covers three categories:
\textit{content removal} (truncation, which drops text without
rewriting it), \textit{structure-preserving rewriting}
(backtranslation), and \textit{partial regeneration} (a fraction of
segments replaced with unwatermarked generations). For each
transformation we report semantic similarity and ROUGE-L against the
untransformed text, alongside the relative change in TokenAcc, IoU,
and EdgeSim relative to clean generation.

According to Table~\ref{tab:robustness}, truncation and weak backtranslation (EN-FR-EN) leave attribution
essentially intact ($\Delta$TokenAcc within $2.4$ points). Stronger
transformations degrade it further: backtranslation through Chinese
costs $5.3$--$5.6$ points on TokenAcc/IoU, and replacing just $10\%$
of segments with unwatermarked text costs $5.6$/$3.6$/$2.4$ points on
TokenAcc/IoU/EdgeSim. These results characterize where the tracing
signal degrades rather than claiming invariance to text
transformation; all transformations tested here are non-adversarial.

\begin{table*}[t]
\centering
\small
\setlength{\tabcolsep}{6pt}
\renewcommand{\arraystretch}{1.15}
\begin{tabular}{ll cc ccc}
\toprule
\textbf{Category} & \textbf{Transformation} & \textbf{SemSim} & \textbf{ROUGE-L}
& $\Delta$\textbf{TokenAcc} & $\Delta$\textbf{IoU} & $\Delta$\textbf{EdgeSim} \\
\midrule
\textit{Clean} & No transformation & --- & --- & $\pm0.0$ & $\pm0.0$ & $\pm0.0$ \\
\midrule
\multirow{2}{*}{\shortstack[l]{Content removal\\\scriptsize(text not rewritten)}}
 & Truncation (90\% kept) & 0.9871 & 0.9444 & $-0.3$ & $-0.8$ & $+2.3$ \\
 & Truncation (75\% kept) & 0.9734 & 0.8542 & $-0.6$ & $-1.5$ & $+2.1$ \\
\midrule
\multirow{2}{*}{\shortstack[l]{Structure-preserving\\rewriting}}
 & Backtranslation (EN-FR-EN) & 0.9251 & 0.7131 & $-2.4$ & $-4.4$ & $+0.8$ \\
 & Backtranslation (EN-ZH-EN) & 0.9156 & 0.6375 & $-5.3$ & $-5.6$ & $-3.3$ \\
\midrule
Partial regeneration\textsuperscript{$\dagger$}
 & 10\% replaced & 0.9741 & 0.9118 & $-5.6$ & $-3.6$ & $-2.4$ \\
\bottomrule
\end{tabular}
\caption{\textbf{Sensitivity of IET to post-generation text transformations}
(\texttt{chain\_4}). We characterize where the tracing signal degrades rather
than claiming invariance to text transformation. $\Delta$ columns give the
relative change (\%) in the Section~\ref{sec:metrics} metrics w.r.t.\ clean
generation; ROUGE-L is measured against the untransformed text. All
transformations are non-adversarial.}
\label{tab:transform_sensitivity}
\label{tab:robustness}
\end{table*}

\section{Ablation: Perturbation Strength $\kappa$ Sweep}
\label{sec:kappa-sweep}

\begin{figure}[t]
  \centering
  \includegraphics[width=0.85\linewidth]{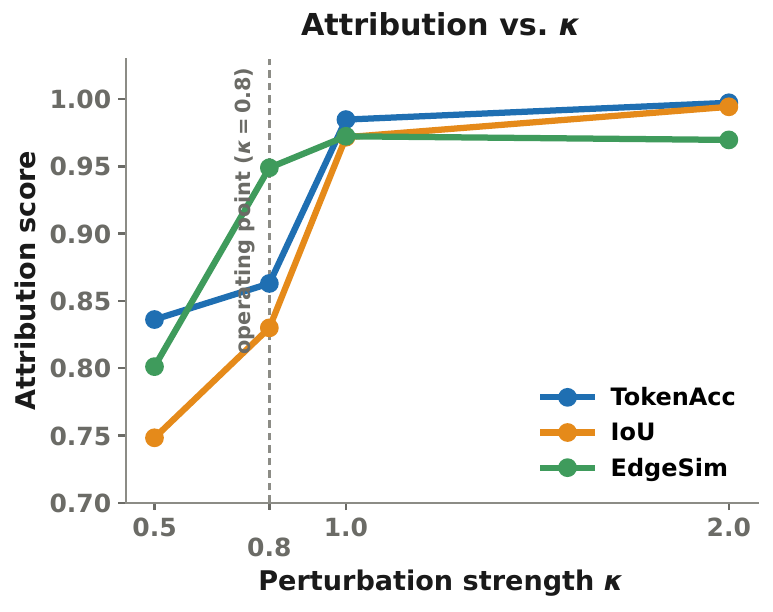}
  \caption{\textbf{Attribution accuracy as a function of perturbation
  strength $\kappa$.} TokenAcc, IoU, and EdgeSim}
  \label{fig:kappa-sweep}
\end{figure}

In Figure~\ref{fig:kappa-sweep}, we sweep the watermark perturbation magnitude $\kappa \in \{0.5, 0.8,
1.0, 2.0\}$ on \texttt{chain\_4}, reporting TokenAcc, IoU, and EdgeSim.
Figure~\ref{fig:kappa-sweep} shows all three rising with $\kappa$. We
select $\kappa=0.8$ (dashed line) as our operating point to balance
attribution accuracy against task performance on normal generation.

\section{Ablation: Sliding-Window Width $w$ Sweep}
\label{sec:w-sweep}

\begin{figure}[t]
  \centering
  \includegraphics[width=0.85\linewidth]{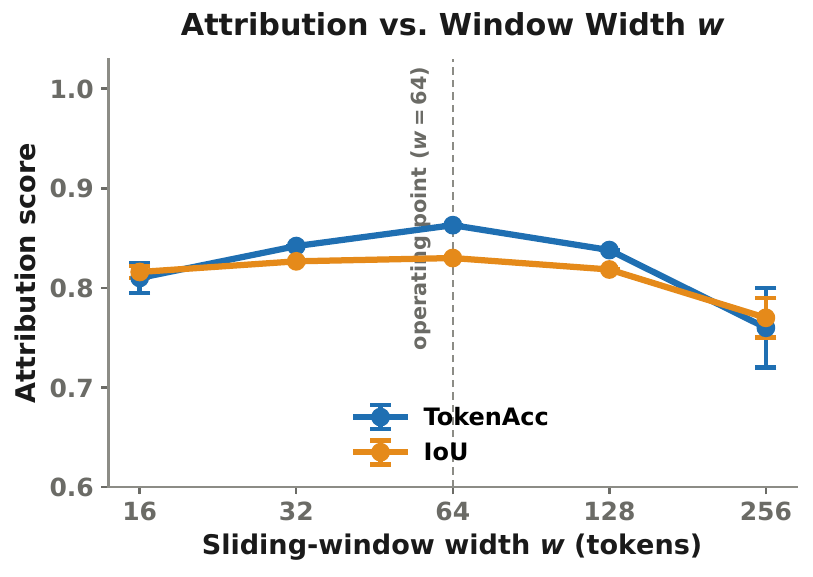}
  \caption{\textbf{Attribution accuracy as a function of sliding-window
  width $w$.}}
  \label{fig:w-sweep}
\end{figure}

In Figure~\ref{fig:w-sweep}, we sweep the sliding-window width $w \in \{16, 32, 64, 128, 256\}$
(Eq.~7) on \texttt{chain\_4}, reporting TokenAcc and IoU from two
independent runs at each width; where the runs agree exactly we
report the measured value, and where they diverge ($w=16$, $w=256$)
we report a central estimate with its spread.

Figure~\ref{fig:w-sweep} shows both metrics rising from $w=16$
($0.810\pm0.015$ / $0.816\pm0.006$) to a peak at $w=64$ ($0.863$ /
$0.830$), then declining through $w=128$ ($0.838$ / $0.8185$) to
$w=256$, where accuracy drops and the two runs disagree widely
(TokenAcc $0.767$ vs.\ $0.639$; IoU $0.792$ vs.\ $0.777$; summarized
as $0.72$--$0.80$ and $0.75$--$0.79$).

We select $w=64$ (dashed line) as our operating point, as it sits at
the peak of both curves. Nearby widths ($32$, $128$) retain most of
the achievable accuracy, so the method is not brittle to this choice,
but both metrics degrade toward $w=16$ and $w=256$.

\section{Robustness to Spurious Registered Keys}
\label{app:spurious-keys}

In decentralized audits the verifier's registry may hold more keys than the
agents active in a given trace, raising the concern that extra, non-contributing
(\emph{spurious}) keys inflate false attributions. Under the chain setting, we find attribution is effectively
unchanged even when the candidate set is more than tripled: see Table~\ref{tab:spurious-keys}. This is
expected, since a non-generating key produces roughly zero-mean scores while the
true key accumulates a consistent positive bias over the span, so as evidence
aggregates spurious argmax collisions become exceedingly unlikely. Attribution is
thus governed by the true signal rather than the size of the registry.

\begin{table}[h]
\centering
\small
\begin{tabular}{@{}cccc@{}}
\toprule
True $K$ & Candidate Keys & TokenAcc & $\Delta$TokenAcc \\
\midrule
4 & 4  & 0.863 & --- \\
4 & 8  & 0.861 & $-0.002$ \\
4 & 12 & 0.860 & $-0.003$ \\
\midrule
6 & 6  & 0.829 & --- \\
6 & 8  & 0.829 & $0.000$ \\
6 & 14 & 0.827 & $-0.002$ \\
\bottomrule
\end{tabular}
\caption{Attribution remains stable as the verifier's candidate-key set grows
beyond the number of active agents (chain setting, $\kappa=0.8$, same pipeline as
Table 1). $\Delta$TokenAcc is measured relative to the matched-key
setting (Candidate Keys $=$ True $K$).}
\label{tab:spurious-keys}
\end{table}

\section{The Usage of Large Language Models (LLMs)}
LLMs were used only occasionally to help polish the writing (e.g., wording suggestions, grammar fixes, and spelling corrections). All technical ideas, experimental designs, analyses, conclusions, and writing were
developed and carried out entirely by the authors. The authors have full responsibility for the final
text.

\end{document}